\documentclass[10pt,twocolumn,letterpaper]{article}

\usepackage{iccv}
\usepackage{times}
\usepackage{epsfig}
\usepackage{graphicx}
\usepackage{amsmath}
\usepackage{fixmath}
\usepackage{amssymb}
\usepackage{color}
\usepackage{multirow}
\usepackage{bm}
\usepackage{caption}
\usepackage{subcaption}
\makeatletter
\@namedef{ver@everyshi.sty}{}
\makeatother
\usepackage{tikz}
\usepackage{pgfplots}

\usepackage[breaklinks=true,bookmarks=false]{hyperref}

\renewcommand{\vec}[1]{\mathbold{#1}}
\newcommand{\mat}[1]{\mathbold{#1}}
\iccvfinalcopy 

\def\iccvPaperID{6577} 

\ificcvfinal\fi

\begin{document}

\title{Class-Incremental Learning for Action Recognition in Videos}

\author{Jaeyoo Park \qquad Minsoo Kang \qquad  Bohyung Han\\
ECE \& ASRI, Seoul National University\\
{\tt\small \{bellos1203,kminsoo,bhhan\}@snu.ac.kr}
}

\maketitle
\ificcvfinal\thispagestyle{empty}\fi



\begin{abstract}
We tackle catastrophic forgetting problem in the context of class-incremental learning for video recognition, which has not been explored actively despite the popularity of continual learning.
Our framework addresses this challenging task by introducing time-channel importance maps and exploiting the importance maps for learning the representations of incoming examples via knowledge distillation.
We also incorporate a regularization scheme in our objective function, which encourages individual features obtained from different time steps in a video to be uncorrelated and eventually improves accuracy by alleviating catastrophic forgetting.
We evaluate the proposed approach on brand-new splits of class-incremental action recognition benchmarks constructed upon the UCF101, HMDB51, and Something-Something V2 datasets, and demonstrate the effectiveness of our algorithm in comparison to the existing continual learning methods that are originally designed for image data.
\end{abstract}



\section{Introduction}
\label{sec:intro}

\begin{figure}[t]
	\centering
	\includegraphics[width=\linewidth]{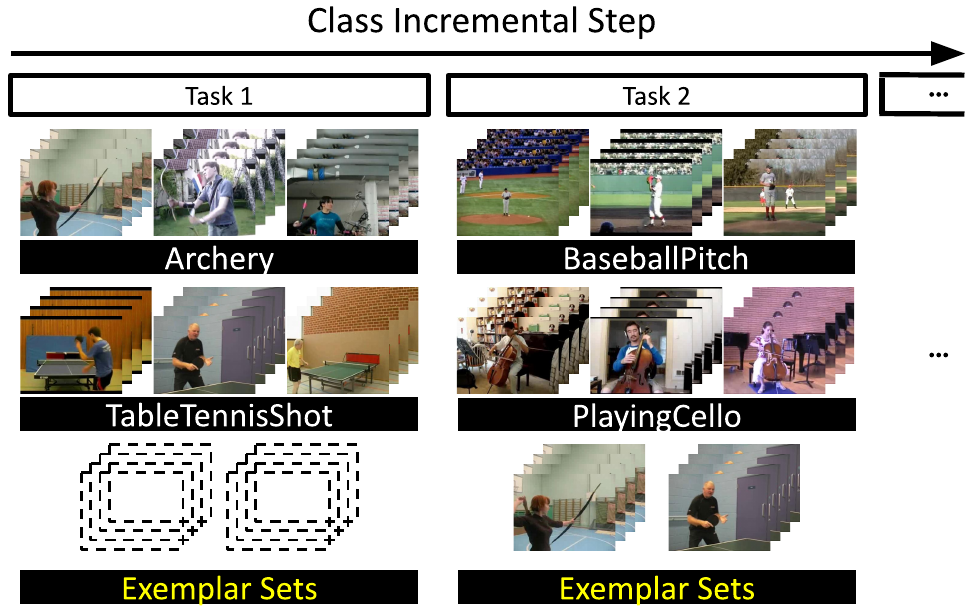}
	\caption{Illustration of class-incremental learning scenario. 
	At each incremental step, the model learns the knowledge of new classes that are disjoint from the classes it has seen so far.
	Simultaneously, the model learns not to forget the knowledge of old classes that are either completely inaccessible or accessible in limited amounts.}
	\vspace{-0.4cm}
\label{fig:class_incremental_learning}
\end{figure}

Human activity recognition in a large-scale video dataset is a crucial step for high-level video understanding, and various approaches have been studied actively in the computer vision community~\cite{carreira2017quo,karpathy2014large,lin2019tsm,tran2015learning,wang2016temporal}. 
If the videos containing unseen classes of actions are presented in a sequential manner, where the examples in the previously observed classes are either inaccessible or accessible in limited amounts, one needs to adapt the current model to the new data without forgetting critical knowledge of the seen examples learned in the past. 
The machine learning paradigm to handle such challenges is called \textit{class-incremental learning}, and
Figure~\ref{fig:class_incremental_learning} illustrates a training data stream for the learning framework. 

While researchers have been studying action recognition problems using deep neural networks~\cite{carreira2017quo,karpathy2014large,lin2019tsm,tran2015learning,wang2016temporal}, continual learning in videos has not been studied actively.
It is natural to claim that video-based recognition tasks are also prone to suffer from catastrophic forgetting~\cite{mccloskey1989catastrophic} for the knowledge learned from training data provided in the past, as in the image domain.
Actually, the catastrophic forgetting problem is particularly problematic in video-learning tasks because deep neural networks with shared parameters are typically applied to multiple segments or frames, resulting in acceleration of the forgetting issue and it is difficult to store many video exemplars in memory to preserve the information about the previous tasks effectively.

\begin{figure}[t]
	\centering
	\begin{subfigure}[t]{1\linewidth}
		\raisebox{-\height}{\includegraphics[width=\linewidth]{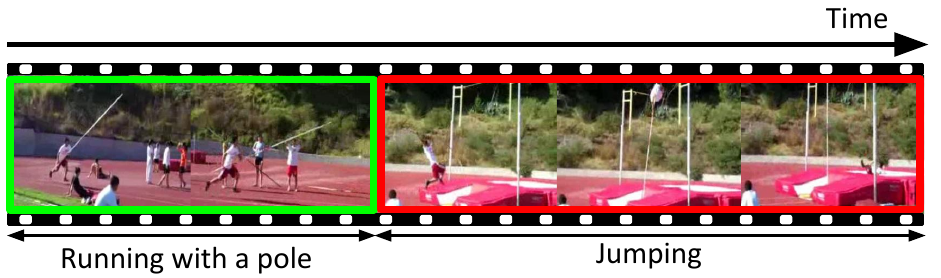}}
		\vspace{-0.2cm}
		\caption{An example of Pole Vault}
	\end{subfigure}
	\begin{subfigure}[t]{1\linewidth}
		\raisebox{-\height}{\includegraphics[width=\linewidth]{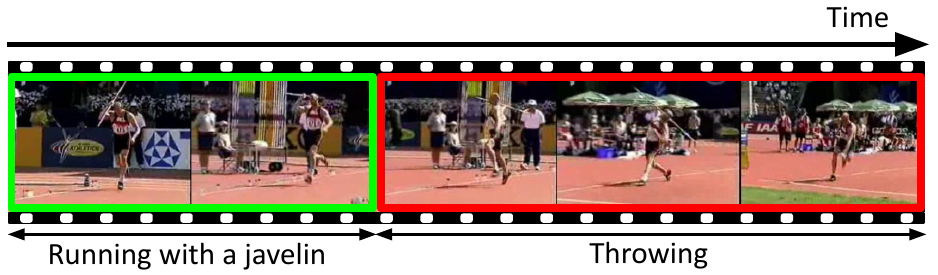}}
		\vspace{-0.2cm}
		\caption{An example of Javelin Throw}
	\end{subfigure}
	\vspace{-0.2cm}
	\caption{Subactions of action instances. 
	Distinctive subactions are key to distinguish one action from another. 
	Our algorithm estimates which channels are important along with the temporal dimension for class-incremental learning.
	}
	\vspace{-0.1cm}
	\label{fig:subaction}
\end{figure}

Despite critical needs for class-incremental learning in the video domain, existing approaches~\cite{castro2018end,  douillard2020podnet, hou2019learning, li2017learning, rebuffi2017icarl, wu2019large} have focused on static images only, which fails to model temporal variations and dynamics across spatial features. 
A single action instance is often composed of multiple subactions and the feature dynamics aligned with the subactions are indeed critical information for action recognition.
For example, Figure~\ref{fig:subaction} demonstrates that both of \textit{Pole Vault} and \textit{Javelin Throw} share a subaction of running with a long stick at the beginning but become distinct by whether the actor jumps or not at the end.
This observation leads to a fundamental question about how to maintain crucial spatio-temporal information within individual videos using limited memory for continual learning.
 

This paper presents a novel framework for class-incremental learning for action recognition based on temporally attentive knowledge distillation.
Our claim is that the representations for individual subactions should be distilled with different weights depending on their relevance and uniqueness to target classes and maintained for better utilization in the future stages.
To realize this idea, we draw our attention to a joint space defined by frames in a video and channels in a feature map, and quantify importance over the space while minimizing feature redundancy across frames.
Specifically, we estimate the importance in the joint space for a video by measuring how much the activation in the space affects classification losses.
The computed importance provides the information about where to attend for knowledge distillation in class-incremental learning scenarios.
Also, to enforce the model to learn more distinctive features across frames, we penalize the redundancy in the features extracted from the sampled frames.

The representations of video data require more computation resources for processing and storing, which makes continual learning in videos more challenging especially when some exemplars for the tasks considered earlier need to be stored in memory. 
So, the proposed class-incremental learning framework employs a frame-based video representation method---Temporal Shift Module (TSM)~\cite{lin2019tsm}, and reduces computational cost for training significantly compared to 3D CNNs based on video volumes~\cite{carreira2017quo,feichtenhofer2020x3d,feichtenhofer2019slowfast,tran2015learning,tran2019video} and their variations~\cite{wang2016temporal,zhou2018temporal}.


The contributions of this paper are summarized below:
\begin{itemize}
	\item We introduce an efficient class-incremental learning technique for action recognition in videos by adopting a simple frame-based feature representation method to store exemplars for the tasks learned in the past. \vspace{-0.1cm}
	
	\item Our algorithm estimates time-channel importances and distills knowledge with the importance weight while encouraging the diversity of the features in each frame for regularization and enhance the performance of our target model. \vspace{-0.1cm}
	
	\item The proposed approach presents remarkable accuracy gains on the multiple standard action recognition benchmarks with brand-new splits compared to the existing methods designed in the image domain.

\end{itemize}

Our paper is organized as follows.
We first discuss related works about continual learning in Section~\ref{sec:related}.
Section~\ref{sec:method} describes the proposed class-incremental learning approach in the context of action recognition.
We present experimental results on the standard action recognition datasets with new splits for continual learning in Section~\ref{sec:experiments}, and make the conclusion in Section~\ref{sec:conclusion}.


\section{Related Works}
\label{sec:related}

This section reviews existing algorithms related to class-incremental learning. 
Most of the researches about continual learning deal with image classification problems only, so we also discuss the approaches in other tasks.

\subsection{Class-Incremental Learning}
Existing class-incremental learning approaches alleviate catastrophic forgetting via the following four techniques: 
1) parameter regularization, 2) knowledge distillation, 3) rehearsal, and 4) bias correction.

\vspace{-0.2cm}
\paragraph{Parameter regularization}
The methods in this category~\cite{aljundi2018memory,kirkpatrick2017overcoming,zenke2017continual} estimate the importance of individual model parameters and exploit the information for model adaptation.
Specifically, the learning algorithm attempts to preserve parameters with high weights while allowing unimportant ones to be flexible for update.
The criteria to determine model elasticity on new tasks include Fisher information matrix~\cite{kirkpatrick2017overcoming}, path integral along parameter trajectory~\cite{zenke2017continual}, and changes in the output vectors~\cite{aljundi2018memory}.
However, these approaches empirically present poor generalization performance in class incremental learning scenarios as reported in \cite{hsu2018re, van2019three}.

\vspace{-0.2cm}
\paragraph{Knowledge distillation}
The approaches based on knowledge distillation~\cite{hinton2014, romero2014fitnets, zagoruyko2016paying} encourage a model to learn new tasks
while mimicking the representations of the old model trained for the previous tasks without their training data. 
To this end, new models attempt to preserve the representations of examples by matching the outputs from the sigmoid functions~\cite{li2017learning, rebuffi2017icarl, wu2019large}, the softmax function with temperature scaling~\cite{castro2018end}, and the $\ell_2$-normalizations~\cite{hou2019learning}.
In addition, LwM~\cite{dhar2019learning} further minimizes the difference of the attention maps obtained from the gradients of the highest score labels. 
PODNet~\cite{douillard2020podnet} preserves the relaxed representations obtained by applying the sum pooling along the width and height dimensions to the original intermediate feature maps and controlling the balance between the previous knowledge and the new information.

\vspace{-0.2cm}
\paragraph{Rehearsal}
Rehearsal-based methods store a limited number of representative examples or replay old ones using generative models while training new tasks.
Incremental Classifier Representation Learning (iCaRL)~\cite{rebuffi2017icarl} keeps a small number of samples per class to approximate the class centroid and makes predictions based on the nearest class mean classifiers.
On the other hand, pseudo-rehearsal techniques~\cite{ostapenko2019learning,shin2017continual} generate samples in the previously observed classes using generative adversarial networks (GANs)~\cite{goodfellow2014generative,odena2017conditional}.
However, generating videos is too challenging to be used for class-incremental learning.

\vspace{-0.2cm}
\paragraph{Bias correction}
The trained models by class-incremental learning algorithms turn out to prefer new classes partly due to the class imbalance problem, and some approaches~\cite{hou2019learning,wu2019large} aim to cope with this issue.
Bias Correction (BiC)~\cite{wu2019large} corrects bias using additional scale and shift parameters for affine transformations of the logits for new classes.  
Zhao~\etal~\cite{zhao2020maintaining} rescale the weight vectors for the new classes by matching the average norm of the old weight vectors.

\subsection{Class-Incremental Learning in Other Domains}
Although class-incremental learning has been studied for image classification, the research is also active for other applications, including person re-identification~\cite{Wu2021Lifelongperson}, 3D object classification~\cite{Dong2021I3DOL}, object detection~\cite{shmelkov2017incremental}, and semantic segmentation~\cite{michieli2019incremental}.
Continual learning in the video domain is rare~\cite{mu2020ilgaco,wang2020catnet}.
Despite remarkable technical advances in action recognition, catastrophic forgetting problem has not been explored actively yet.
An existing approach~\cite{wang2020catnet} is limited to applying the iCaRL~\cite{rebuffi2017icarl} based on a two-stream 3D convolutional neural network in a straightforward manner.
On the other hand, our approach is based on knowledge distillation similar to \cite{dhar2019learning,douillard2020podnet} and exploits an attention method over a time-channel space intuitively to facilitate action recognition in a class-incremental learning scenario.

\subsection{Action recognition}
With the great success of deep learning, various action recognition methods based on convolutional neural networks have been studied~\cite{carreira2017quo,karpathy2014large,lin2019tsm,tran2015learning,wang2016temporal}. 
The approaches to handle this problem are grouped in two categories: 2D and 3D CNN-based methods.
2D CNN-based techniques~\cite{karpathy2014large,lin2019tsm,wang2016temporal} utilize the standard CNN models~\cite{he2016deep,simonyan2014very} on each frame, and the researchers have explored how to aggregate the information from each time step~\cite{karpathy2014large}. 
For example, \cite{feichtenhofer2016convolutional} studies how to fuse the information from two different modalities, RGB and motion, using 2D CNNs.
Wang \etal~\cite{wang2016temporal} propose a strategy to learn with uniformly divided segments in multiple modalities, \ie RGB difference and warped optical flow.
In \cite{zhou2018temporal}, they learn temporal dependencies across frames by exploring multiple time scales.
Recently, Temporal Shift Module (TSM)~\cite{lin2019tsm} proposes a method to learn temporal information in an efficient way, where the feature representations of adjacent segments interact with each other during forward pass. 

On the other hand, some researchers pay attention to 3D CNN~\cite{carreira2017quo,feichtenhofer2020x3d,feichtenhofer2019slowfast,tran2015learning,tran2019video}, which is a straightforward extension of 2D CNN methods, where 3D convolution filters learn spatio-temporal information jointly.
However, 3D CNN-based models are computationally expensive since they involve a large number of parameters to learn.
Recent approaches handle this issue in diverse ways, for example, by applying group convolutions~\cite{tran2019video}, learning 3D shift operations~\cite{fan2020rubiksnet}, decomposing 3D convolution filters~\cite{tran2018closer}, searching efficient architectures~\cite{feichtenhofer2020x3d}, \etc.
Despite remarkable advances in action recognition, the catastrophic forgetting problem is not yet studied actively.
This work sheds light on this problem with a promising baseline.


\begin{figure*}[t!]
\begin{center}
   \includegraphics[width=\linewidth]{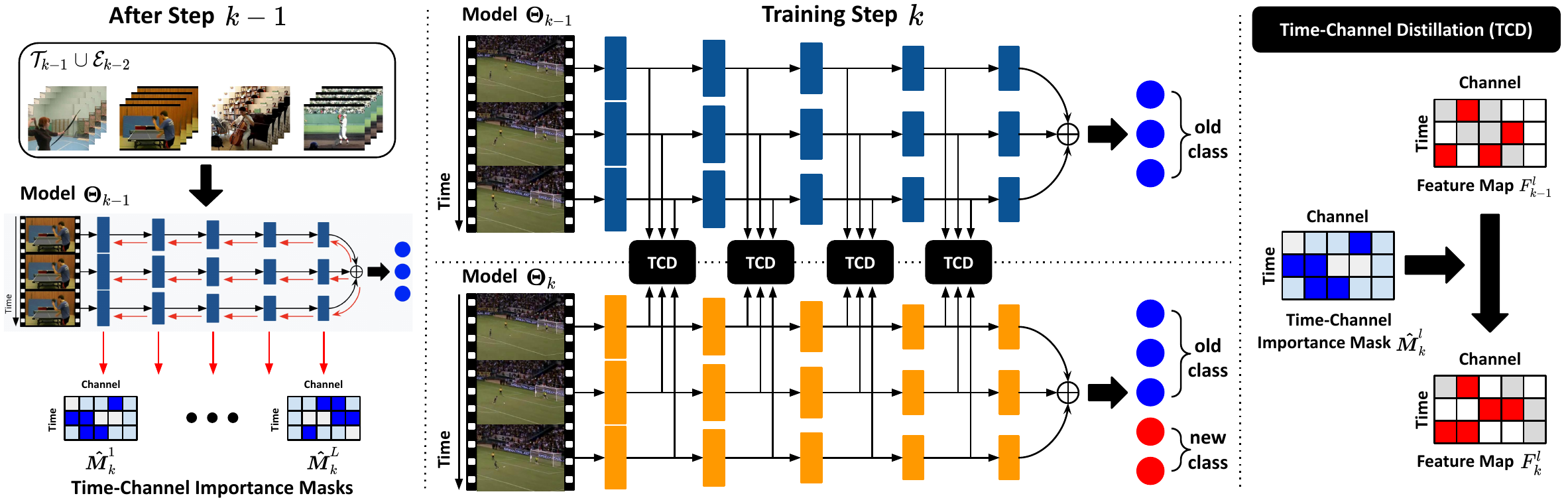}
\end{center}
\vspace{-0.3cm}
   \caption{Illustration of the overall framework. 
   At each incremental step $k$, the current model $\boldsymbol{\Theta}_{k}$ (bottom-center) mimics the representation of the previous model $\boldsymbol{\Theta}_{k-1}$ (top-center).
   The distillation process is enhanced by the time-channel importance mask, which is estimated at step $k-1$ via measuring how each feature affects the final loss.
   Distilling knowledge through the estimated importance map makes the model preserve important representations from the previous step; the less important representations are suppressed by the mask and updated flexibly for the new task.}
\label{fig:overall}
\end{figure*}

\section{Method}
\label{sec:method}
This section describes the overall framework of the proposed class-incremental learning algorithm with videos.

\subsection{Problem Formulation}
\label{sec:method_formulation}
The goal of class-incremental learning is to train a unified deep neural network parameterized by $\boldsymbol{\Theta}$ given a sequence of tasks, $\{\mathcal{T}_1, \mathcal{T}_2, \cdots, \mathcal{T}_k, \cdots\}$.
We denote $\mathcal{T}_k$ as a set of videos whose labels belong to the predefined classes in $\mathcal{C}_k$, where $ (\mathcal{C}_1 \cup \cdots \cup \mathcal{C}_{k-1}) \cap \mathcal{C}_k = \emptyset$.
We assume that we can access a small exemplar set denoted by $\mathcal{E}_k$ such that it is a subset of $\mathcal{T}_{1:k}$, where $\mathcal{T}_{1:k}$ = $\mathcal{T}_1 \cup \cdots \cup \mathcal{T}_k$. 
At each incremental step $k$, a model $\boldsymbol{\Theta}_{k}$ is trained with $\mathcal{T}'_{k}={\mathcal{T}_{k} \cup \mathcal{E}_{k-1}}$.
Then, the performance of the trained model is evaluated on the test examples defined by the union of all the encountered tasks without task boundaries. 


\subsection{Overview}
\label{sec:method_overview}
Given the problem formulation, we follow the standard class-incremental learning protocol based on knowledge distillation, which is similar to the previous works~\cite{douillard2020podnet,hou2019learning,li2017learning, rebuffi2017icarl}.
At the $k^\text{th}$ incremental step, a set of model parameters, $\boldsymbol{\Theta}_k$, is learned to mimic the feature representations given by the previous model with $\boldsymbol{\Theta}_{k-1}$ while learning new classes. 
Our goal is to estimate desirable attention over a combination of time and channel dimensions for knowledge distillation. 

Figure~\ref{fig:overall} illustrates the overall framework of our approach.
Given an input video $x \in \mathcal{T}'_{k}$, we first divide the video into $T$ segments, and then feed the segments to the backbone network with $L$ layers.
We adopt TSM~\cite{lin2019tsm} as our backbone model.
Let $\mat{F}^{l}_k \in \mathbb{R}^{T \times C_l \times H_l \times W_l }$ be an intermediate feature with respect to the input $x$ in the $l^\text{th}$ layer of the model $\boldsymbol{\Theta}_k$.
The distillation between $\mat{F}_k^l$ and $\mat{F}_{k-1}^l$ is weighted by the importance mask $\hat{\mat{M}}_{k}^l\in \mathbb{R}^{T \times C_l}$, which is the key component of our framework.
The importance mask $\hat{\mat{M}}_{k}^l$ represents the information about which feature maps along time or channel dimensions are important to preserve knowledge for the past tasks. 

After each incremental step, we select a set of video instances in $\mathcal{T}_k$ to update the exemplar memories from $\mathcal{E}_{k-1}$ to $\mathcal{E}_{k}$ by the herding strategy~\cite{rebuffi2017icarl}.
Then, we fine-tune the final classification layer while freezing other layers using $\mathcal{E}_{k}$, which has balanced data among the observed classes as discussed in \cite{douillard2020podnet,hou2019learning}.

\subsection{Time-Channel Importance}
\label{sec:method_importance}
We focus on designing the importance mask so that it provides each feature map of a frame with the information about which channels should be preserved against the catastrophic forgetting problem. 
Specifically, we aim to keep the important feature maps whose update is prone to increase the final loss, and make the unimportant ones flexible for future tasks.
To this end, we compute the importance of channel $c$ at time step $t$ for incremental step $k$ in the $l^{\text{th}}$ layer, which is denoted by $\mat{M}_{k,t,c}^l$, as
%
\begin{equation}\label{eq:importance_map}
\begin{aligned}
\mat{M}_{k,t,c}^l = \mathbb{E}_{(x,y)\sim\mathcal{T}_{1:k-1}}\|\nabla_{\mat{F}_{k-1,t,c}^l}\mathcal{L}_{\text{cls}}^{k-1}(x,y)\|^2_F,
\end{aligned}
\end{equation}
where $\| \cdot \|_F$ and $\mathcal{L}_{\text{cls}}^{k-1}(x,y)$ denote the Frobenius norm and the classification error of the trained model parametrized by $\boldsymbol{\Theta_{k-1}}$ for the input video $x$ and its label $y$.
Since the perturbation in the feature map with the higher Frobenius norm of the gradient may result in larger increase of the final loss when feature maps are equally important, the importance mask $\mat{M}_{k,t,c}^l$ can be regarded as the sensitivity to the final loss.
Thus, at the end of each incremental step, \ie, $\mathcal{T}_{k-1}$, we sequentially update the important mask $\mat{M}_{k}^l$ for training the new model $\boldsymbol{\Theta}_k$.

However, due to the restriction of the class-incremental learning, we can access limited samples for $\mathcal{T}_{1:k-2}$ using the exemplar sets $\mathcal{E}_{k-2}$, which makes it difficult to compute $\mat{M}_{k,t,c}^l$.
Hence, we approximate $\mat{M}_{k,t,c}^l$ to $\tilde{\mat{M}}_{k,t,c}^l$ by taking the expectation of the Frobenius norm over the accessible samples within $\mathcal{E}_{k-2}$ and $\mathcal{T}_{k-1}$. 
Furthermore, we normalize $\tilde{\mat{M}}_{k,t,c}^l$ to make the importance across layers have a similar scale as follows:
\begin{equation}\label{eq:normalize_maps}
\begin{aligned}
\hat{\mat{M}}_{k,t,c}^l  = \frac{\tilde{\mat{M}}_{k,t,c}^l }{\frac{1}{TC_l}\sum_{t=1}^T\sum_{c=1}^{C_l}\tilde{\mat{M}}_{k,t,c}^l }.
\end{aligned}
\end{equation}
%
%
Finally, we define the proposed distillation loss for the intermediate features in the new model based on the importance map as
\begin{equation}\label{eq:dist_feat}
\begin{aligned}
\mathcal{L}_\text{dist}^k =\sum_{l=1}^{L}\sum_{t=1}^{T}\sum_{c=1}^{C_l}\hat{\mat{M}}_{k,t,c}^l \|\mat{F}_{k, t,c}^l - \mat{F}_{k-1, t,c}^l \|_F^2.
\end{aligned}
\end{equation}
%
The proposed distillation loss constrains the model divergence of the sensitive feature maps not to forget the previously learned knowledge and makes the uncritical ones flexible to learn new classes. 

\subsection{Orthogonality between Frames}
\label{sec:method_orthogonality}
To further improve the effectiveness of the proposed knowledge distillation strategy, we adopt an additional regularization term inspired by~\cite{lin2017structured}, which enforces individual features extracted from different time steps in a video to be mutually independent.
The corresponding loss constrains the features at individual time steps to be orthogonal, which also makes the estimation of the importance map more distinctive.
The orthogonality loss is defined by
\begin{equation}\label{eq:orthogonality}
\begin{aligned}
\mathcal{L}_\text{ortho}^k = \sum_{l=1}^L\sum_{c=1}^{C_l}\|\mat{I}_T-\mat{F'}_{k,:,c}^l (\mat{F'}_{k,:,c}^l)^{\top}\|_F^2,
\end{aligned}
\end{equation}
where $\mat{I}_T \in \mathbb{R}^{T \times T}$ is an identity matrix and $\mat{F'}_{k,:,c}^l$ is given by concatenating a reshaped tensor of $\ell_2$-normalized $\mat{F}_{k,t,c}^l$ along time axis $t$ and constructing a $T\times H_lW_l$ matrix.

The orthogonality constraint is useful in continual learning scenarios since we often need to update model parameters based on limited observations of old data but a large number of examples in new tasks.
Since such a challenging situation leads to unwanted representation changes of the exemplars representing previous tasks, the minimization of correlation between the representations of individual frames would help alleviate the feature drift issue.

\subsection{Training Objective}
\label{sec:method_objective}
The formal definition of the final objective function $\mathcal{L}_{\text{final}}^k$ at incremental step $k$ is given by
\begin{equation}
\mathcal{L}_{\text{final}}^k = \mathcal{L}_\text{cls}^k + \alpha\mathcal{L}_\text{dist}^k +  \beta \mathcal{L}_\text{ortho}^k,
\label{eq:loss_definition}
\end{equation}
where $\alpha$ and $\beta$ are the weights for the balance between the terms. 
For the classification loss $\mathcal{L}_{\text{cls}}^k$, we adopt NCA loss~\cite{movshovitz2017no} computed from the \textit{Local Similarity Classifier} (LSC) following~\cite{douillard2020podnet}. 

\subsection{Exemplar Selection}
\label{sec:method_exemplar}
After each incremental step $k$, we sample the most representative instances from $\mathcal{T}_k$ to construct $\mathcal{E}_k$, for future use.
We follow the herding strategy proposed by~\cite{rebuffi2017icarl}, for which the feature representations for all video samples are extracted from $\mathcal{T}_k$ and the class-wise mean features are computed.
Then we iteratively select the instances for each of the classes until the number of selected exemplars reaches a predefined memory budget.
At each iteration, we choose the exemplar that makes the mean of exemplars become closest to the real class-mean representation.

When we store videos as exemplars, we can further reduce the memory requirement by sampling frames within the video since a single video contains many repetitive and redundant frames.
For each video, we have three options: storing an entire video, sampling frames randomly, or selecting frames with a uniform time interval.
Among the three strategies, we choose the last one, storing the uniformly sampled $T$ frames per video, which meets the input specification of our backbone model, TSM~\cite{lin2019tsm}.
We further discuss this sampling strategy in Section~\ref{sub:memory}.


\begin{table*}[ht!]
	\caption{Class-incremental action recognition performance on UCF101 and HMDB51 of the tested algorithms.
	The proposed method, TCD, achieves the best performance in all the experimental settings.
	NME scores for the methods without exemplars cannot be reported while iCaRL reports NME scores only since iCaRL employs NME for classification.
	The bold-faced numbers indicate the best performance.}
	\label{tab:main_table}
	\vspace{-0.5cm}
	\begin{center}
		\scalebox{0.9}{
			\renewcommand{\arraystretch}{1.2}
			\setlength\tabcolsep{7pt}
			\begin{tabular}{l|cccccc|cccc}
& \multicolumn{6}{c|}{UCF101} & \multicolumn{4}{c}{HMDB51} \\  \hline			 
\multicolumn{1}{l|}{Num. of classes}              &  \multicolumn{2}{c}{10 $\times$ 5 stages}   &  \multicolumn{2}{c}{5 $\times$ 10 stages} &  \multicolumn{2}{c|}{2 $\times$ 25 stages}   & \multicolumn{2}{c}{5 $\times$ 5 stages} &  \multicolumn{2}{c}{1 $\times$ 25 stages}   \\ 
Classifier &   CNN  & NME      & CNN         & NME      & CNN        & NME       & CNN       & NME    & CNN        & NME \\ \hline
				 Fine-tuning                                                                &   24.97 &  ---          & 13.45       &    ---        &  5.78     &  ---          &  16.82     &  ---       &  4.83       & ---                          \\
				 LwFMC~\cite{li2017learning, rebuffi2017icarl}         &  42.14  &  ---          & 25.59     &    ---        &  11.68     &  ---          &   26.82    &  ---       &  16.49         & ---                    \\
				 LwM~\cite{dhar2019learning}                                   &  43.39  &  ---         &  26.07     &   ---           &  12.08       &  ---          &   26.97    &  ---       &  16.50        & ---                    \\
				 iCaRL~\cite{rebuffi2017icarl}                                    &  ---  &  65.34    &   ---     &   64.51     &  ---    &   58.73     &  ---     & 40.09   &   ---       &    33.77                 \\
				 UCIR~\cite{hou2019learning}                                   &   74.31  & 74.09    &  70.42      &   70.50    &  63.22      &  64.00    &  44.90    &  46.53  &  37.04      &  37.15    \\
				 PODNet~\cite{douillard2020podnet}                        &   73.26  &  74.37    &  71.58     &   73.75     &  70.28      &  71.87   & 44.32     &   48.78  &  38.76    & 46.62   \\ \hline 
				 TCD (Ours)                                                                         &    \textbf{74.89} &  \textbf{77.16} &  \textbf{73.43}    &  \textbf{75.35}   &  \textbf{72.19}         & \textbf{74.01} &\textbf{45.34} & \textbf{50.36} &  \textbf{40.07} &  \textbf{46.66}          \\ \hline
				 Oracle (Upper Bound)                                                           & 84.15                & 83.37              & 83.96                  & 83.20                 & 83.82 & 83.16 & 55.03 & 55.98 & 54.89 & 55.32
			\end{tabular} 
		}
	\end{center}
	\vspace{-0.3cm}
\end{table*}

\begin{table}[t!]
    \caption{Class-incremental action recognition performance on Something-Something V2.
    The bold-faced numbers indicate the best performance.}
    \label{tab:sthsth_table}
    \vspace{-0.2cm}
    \centering
    \scalebox{0.9}{
		\setlength\tabcolsep{7pt}
		\begin{tabular}{l|cc|cc}		 
\multicolumn{1}{l|}{Num. of classes}              &  \multicolumn{2}{c|}{10 $\times$ 9 stages}   &  \multicolumn{2}{c}{5 $\times$ 18 stages}    \\ 
Classifier &   CNN  & NME      & CNN         & NME       \\ \hline \hline
                                   UCIR~\cite{hou2019learning}                             &   26.84  &  17.98   &  20.69    &  12.57    \\
				 PODNet~\cite{douillard2020podnet}                   &   34.94  &  27.33   &  26.95    &  17.49     \\ \hline  
				 TCD (Ours)                                                           & \textbf{35.78}               & \textbf{28.88}              & \textbf{29.60}                & \textbf{21.63} \\ \hline        
		\end{tabular} 
	}
\end{table}

\section{Experiments}
\label{sec:experiments}
This section presents the experimental results of our algorithm on class-incremental action recognition benchmarks.
We also demonstrate the effectiveness of our framework via several ablation studies.

\subsection{Datasets} 
\label{sub:datasets}
We evaluate the proposed framework on UCF101~\cite{soomro2012ucf}, HMDB51~\cite{kuehne11hmdb} and Something-Something V2~\cite{goyal2017something}, which are the standard datasets for action recognition tasks.
The UCF101 dataset consists of 13.3K videos from 101 classes.
The organizers of UCF101 provide three splits of training and test datasets.
The HMDB51 dataset consists of 6.8K examples from 51 action classes, and also provides three splits for training and test datasets.
We adopt split 1 for both of the datasets to evaluate our approach.
Something-Something V2 dataset is a large-scale motion-sensitive dataset, which contains 169K training and 25K test videos from 174 action classes.
This dataset requires better temporal reasoning than UCF101 and HMDB51.

\subsection{Evaluation Protocol}
\label{sub:evaluation_protocol}
Since the aforementioned datasets are utilized for the class-incremental learning for the first time, we newly design the experimental protocol for the datasets. 
We first shuffle the classes randomly to create a sequence of classes.
Following~\cite{douillard2020podnet,hou2019learning}, we assume that we initially have a trained model with half of the total classes, where the rest of the classes are provided sequentially in each incremental step.
For UCF101, we trained 51 classes in the initial stage, and divided the remaining classes into groups of 10, 5, and 2 classes for class-incremental learning. 
For HMDB51, we learned the initial model using 26 classes and the remaining classes are equally split into 5 and 25 groups. 
After obtaining the initial model with 84 classes for Something-Something V2, we generate groups of 10 and 5 classes.

At each incremental step, we evaluate the model with the test data of all the seen classes until then. 
Following the previous works, we employ two methods for inference, CNN and NME, respectively.
CNN is a standard classification protocol, where the model classifies the data using the trained fully-connected layer.
NME, which is proposed by iCaRL~\cite{rebuffi2017icarl}, compares the feature representation of test data with the mean-of-exemplars.
We report the average of the accuracies aggregated from all of the incremental steps, which is also known as \textit{average incremental accuracy}~\cite{douillard2020podnet,hou2019learning,rebuffi2017icarl}.
Since the order of classes may affect the performance, we ran our experiments using three random class orders\footnote{Random Seeds : 1000, 1993, 2021} and report the average performance.
We set the memory budget for each class to 5 for UCF101 and HMDB51, and 20 for Something-Something V2 unless specified otherwise.

\subsection{Implementation Details}
\label{sub:imp_details}
We construct our framework based on the official implementation of TSM\footnote{https://github.com/mit-han-lab/temporal-shift-module} using the PyTorch library~\cite{paszke2019pytorch}.
We follow data pre-processing protocol of TSM.
For UCF101, we train a ResNet-34 TSM for 50 epochs with a batch size of 32.
For HMDB51 and Something-Something V2, we adopt ResNet-50 models and train for 50 epochs with a batch size of 64.
For all datasets, we use the ImageNet-pretrained weights for initialization.
Note that we do not use the weights pretrained with the Kinetics dataset~\cite{carreira2017quo}, which is common in the action recognition field.
It is inappropriate to evaluate class-incremental learning with the weights pretrained with Kinetics since it shares the class information with UCF101 and HMDB51. 
Thus the pretrained weights already contain the class information of target datasets.
Please refer to the supplementary materials for more implementation details.

\begin{figure*}[h!]
	\centering
	\begin{subfigure}[t]{0.3\textwidth}
		\raisebox{-\height}{\includegraphics[width=\textwidth]{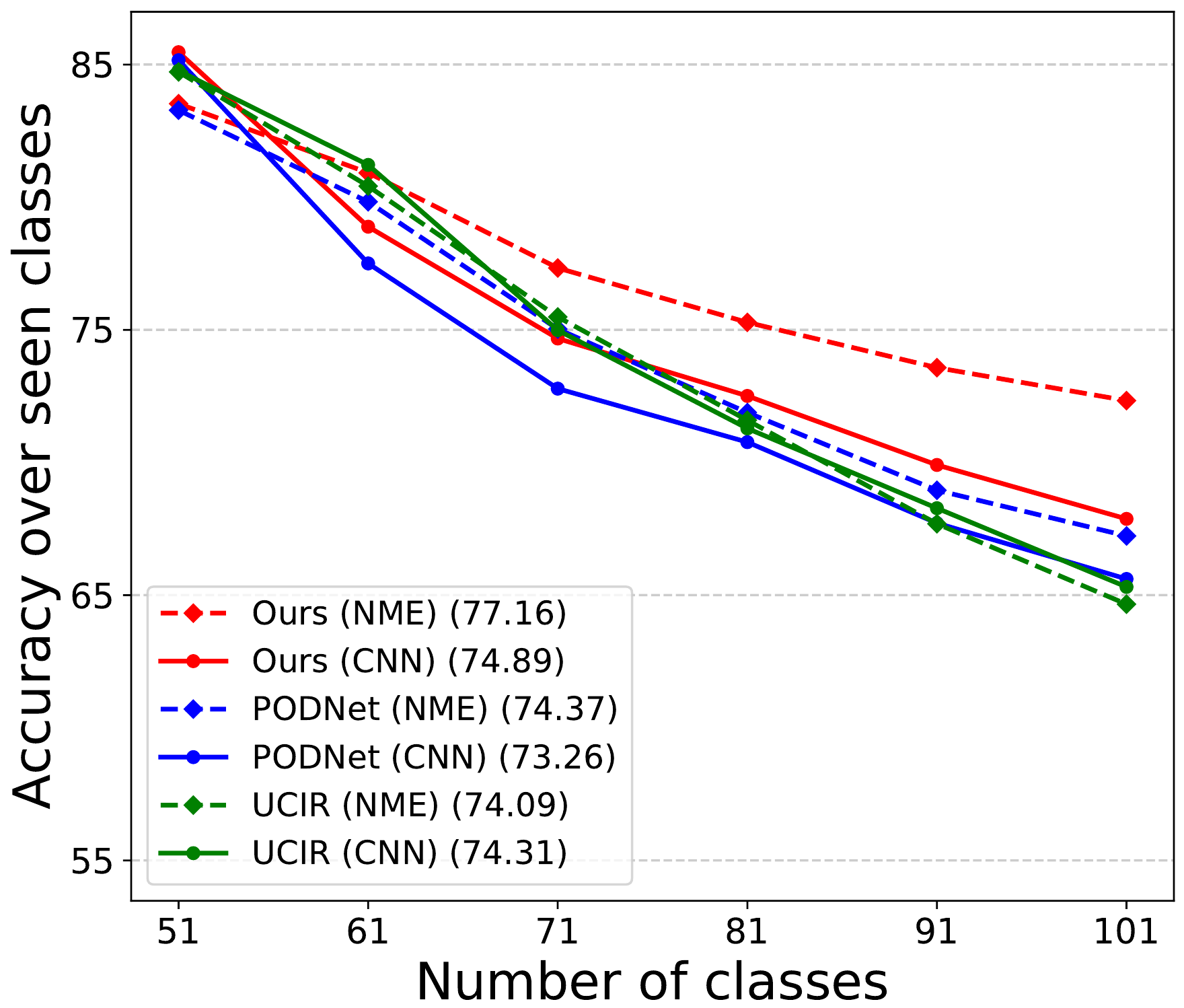}} 
		\subcaption{UCF101 ($10\times5$ stages)}
	\end{subfigure} \hspace{0.2cm}
	\begin{subfigure}[t]{0.3\textwidth}
		\raisebox{-\height}{\includegraphics[width=\textwidth]{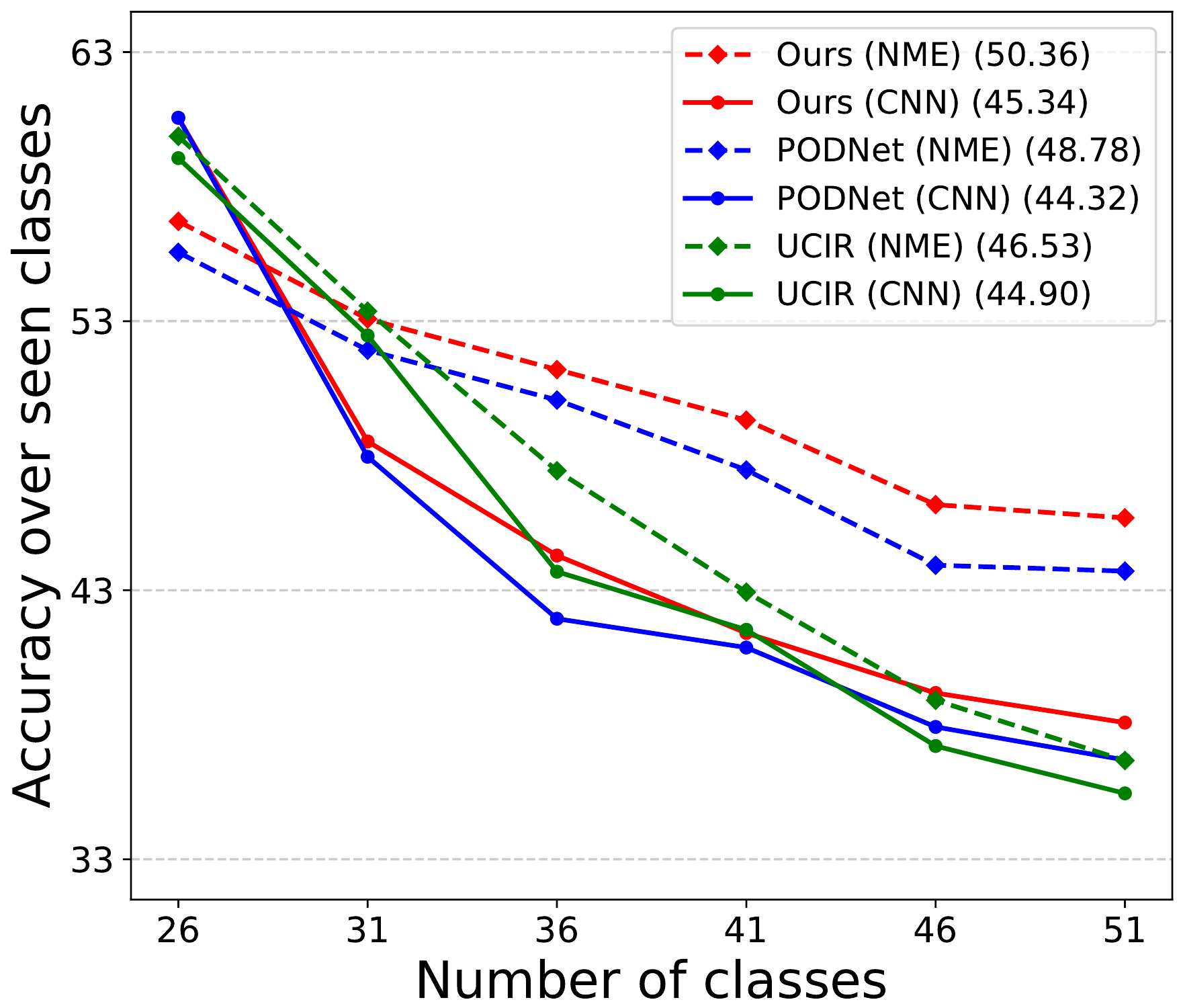}}
		\subcaption{HMDB51 ($5\times5$ stages)}
	\end{subfigure} \hspace{0.3cm}
	\begin{subfigure}[t]{0.3\textwidth}
		\raisebox{-\height}{\includegraphics[width=\textwidth]{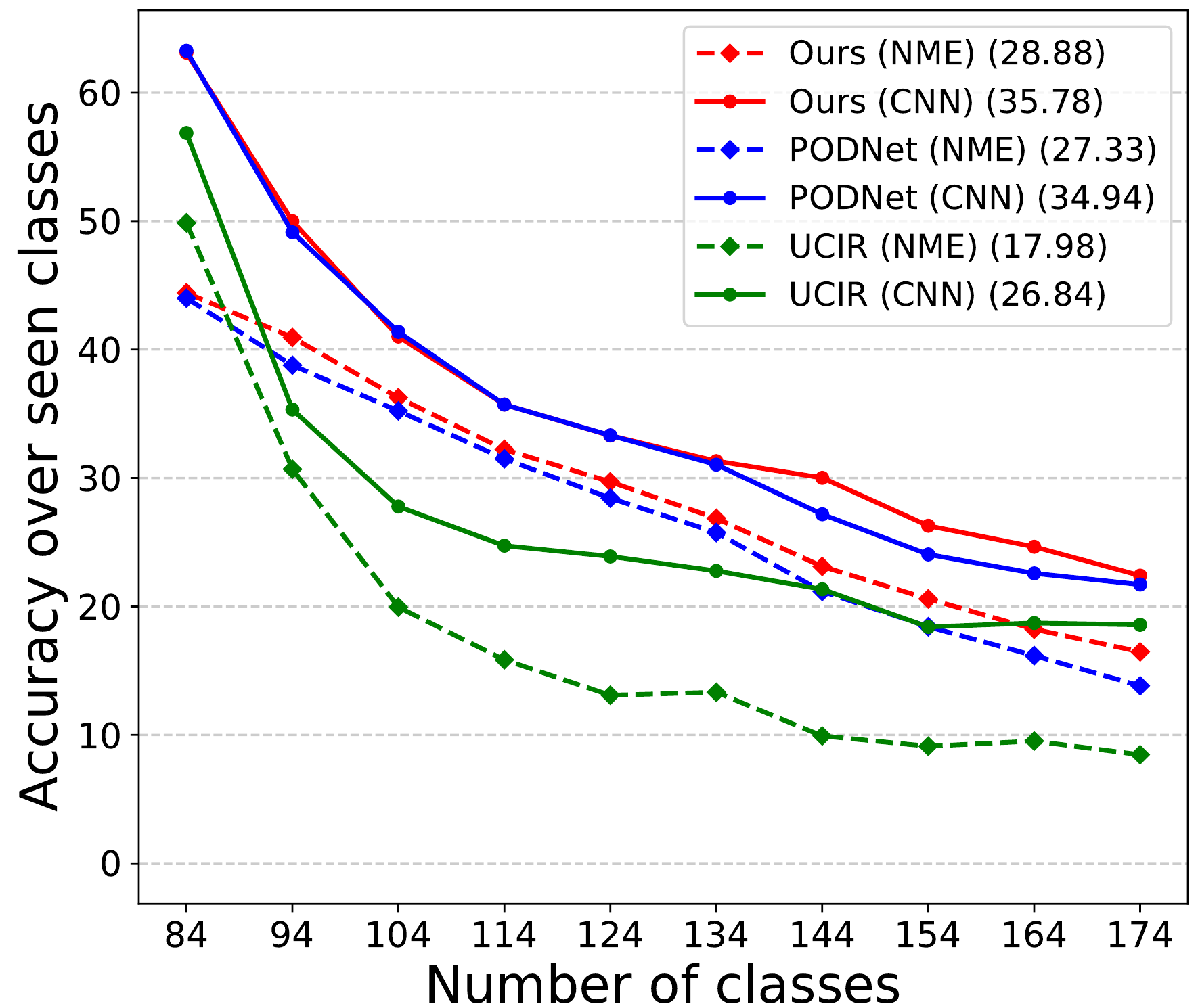}}
		\subcaption{Sth-Sth-V2 ($10\times9$ stages)}
	\end{subfigure}
	\begin{subfigure}[t]{0.3\textwidth}
		\raisebox{-\height}{\includegraphics[width=\textwidth]{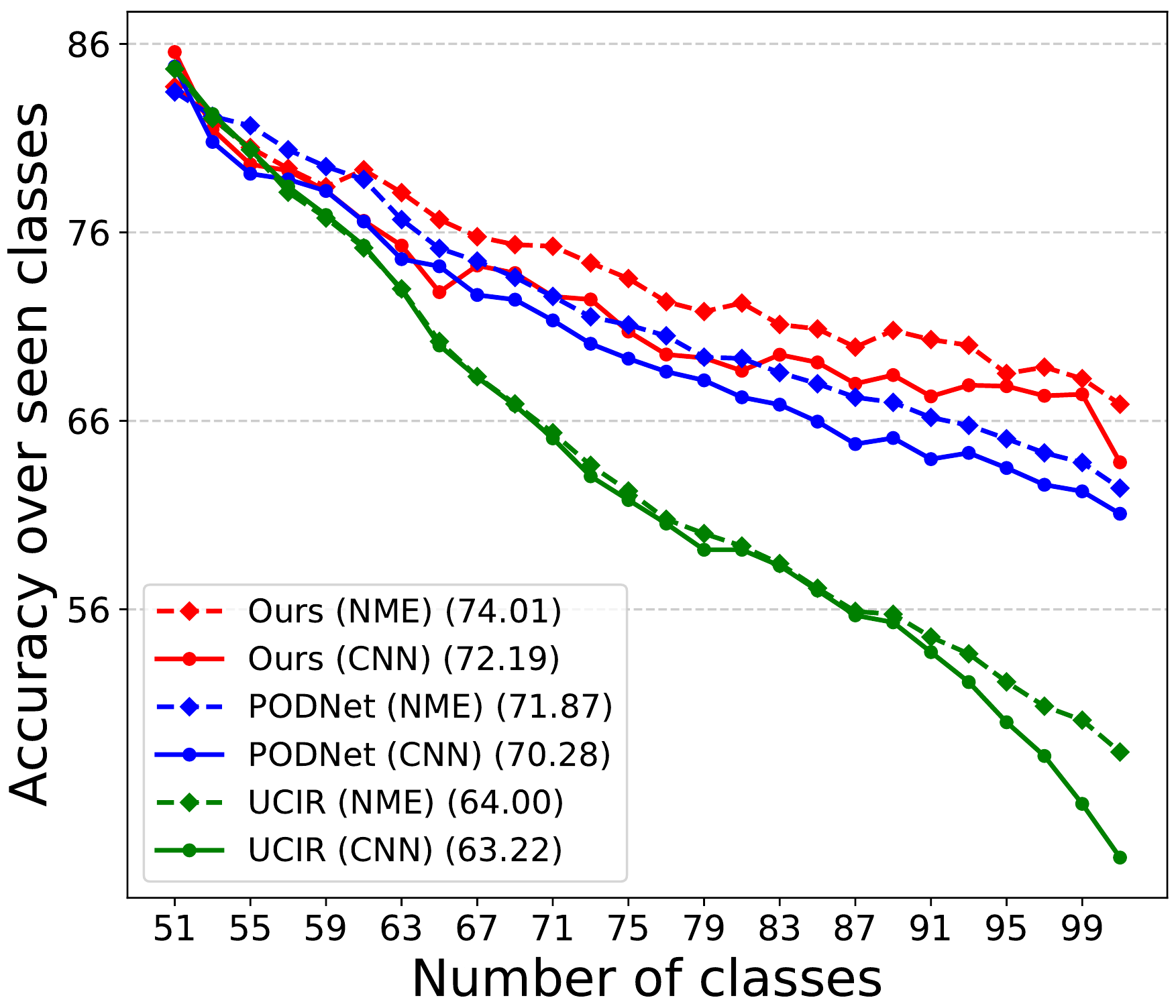}}
		\subcaption{UCF101 ($2\times25$ stages)}
	\end{subfigure} \hspace{0.2cm}
	\begin{subfigure}[t]{0.3\textwidth}
	    \raisebox{-\height}{
	    \includegraphics[width=\textwidth]{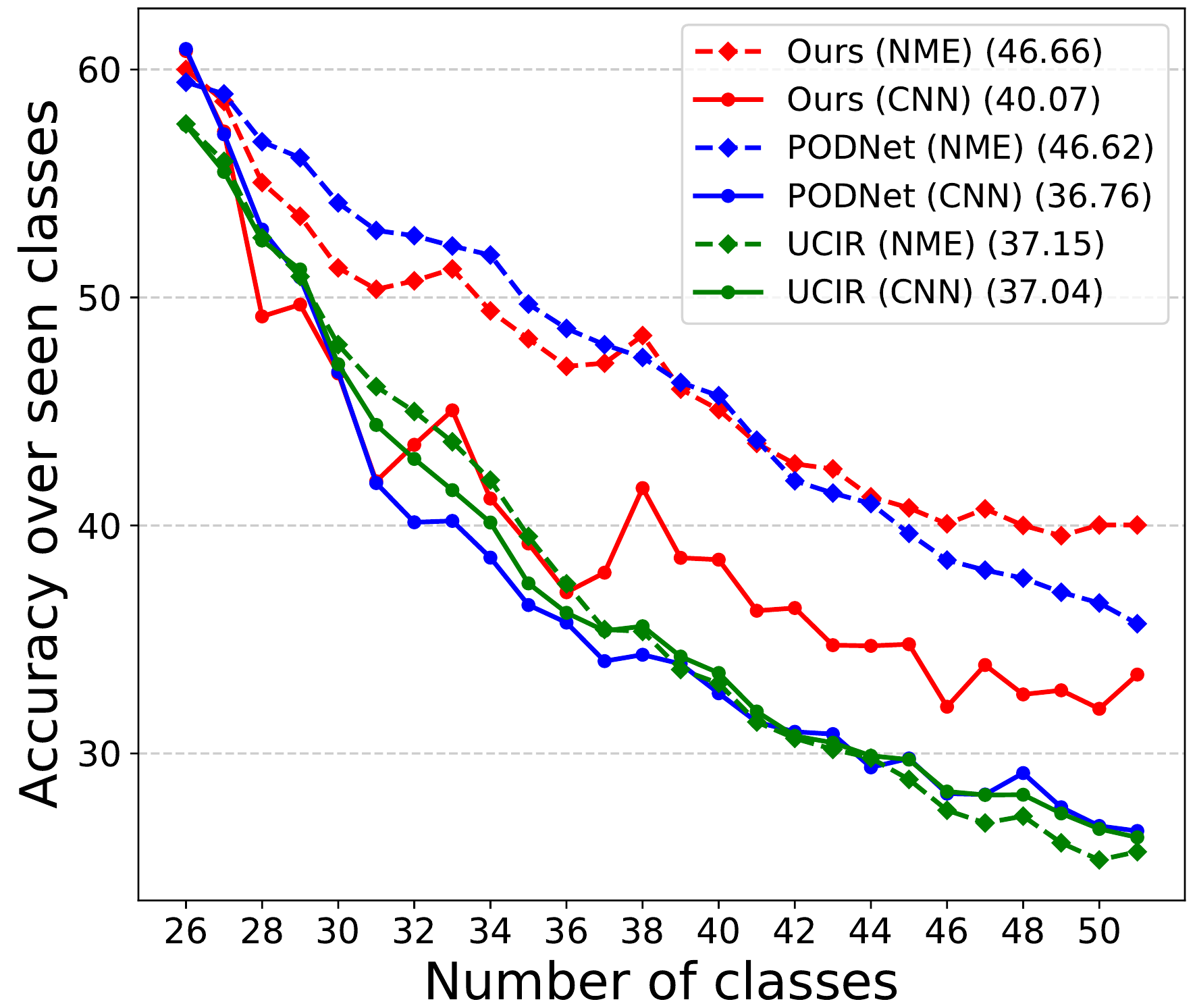}}
		\subcaption{HMDB51 ($1\times25$ stages)}
	\end{subfigure} \hspace{0.3cm}
	\begin{subfigure}[t]{0.3\textwidth}
		\raisebox{-\height}{\includegraphics[width=\textwidth]{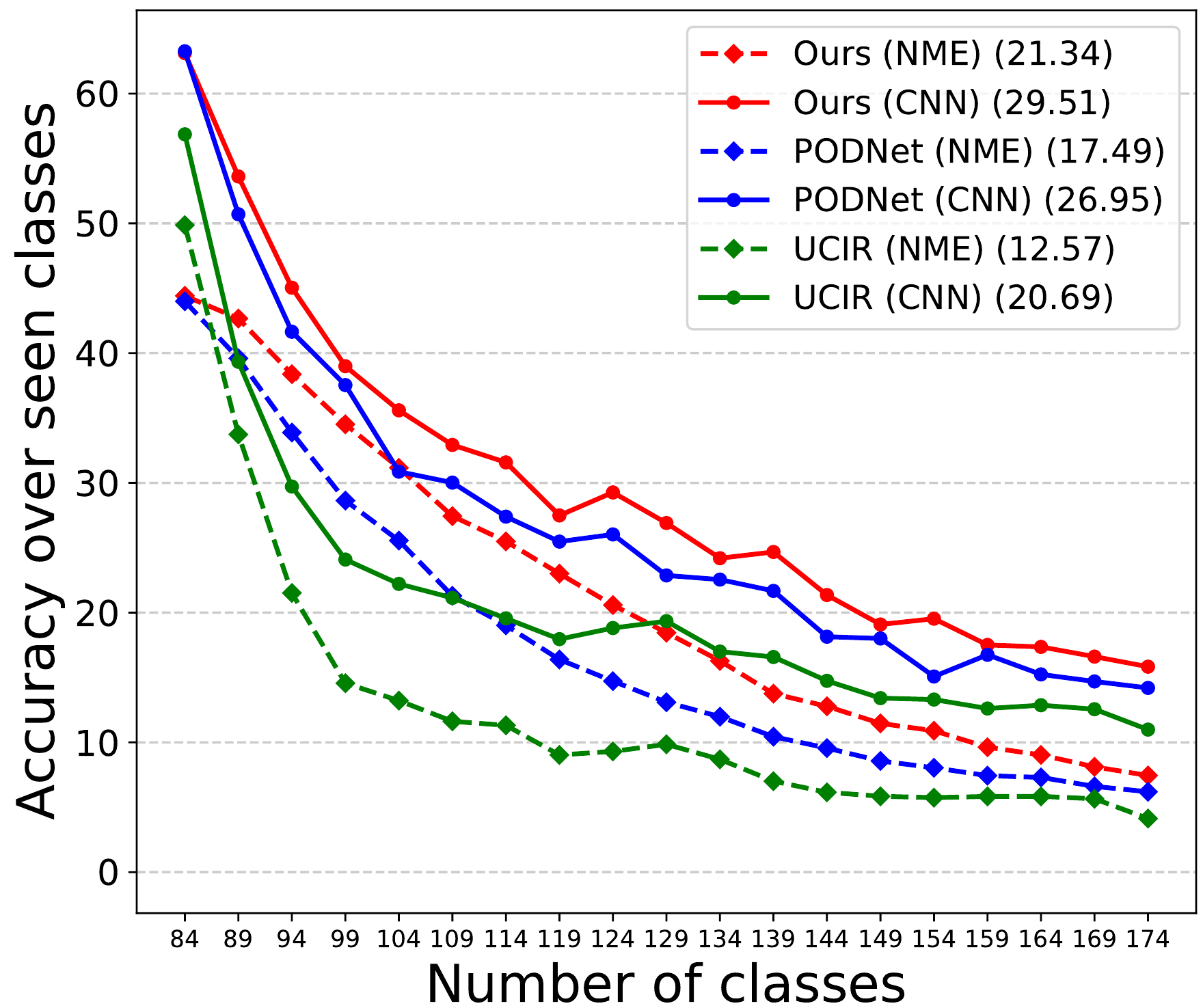}}
		\subcaption{Sth-Sth-V2 ($5\times18$ stages)}
	\end{subfigure}
	\vspace{-0.2cm}
	\caption{Plots for accuracy on UCF101, HMDB51 and Something-Something V2 along with the incremental steps.}
	\label{fig:acc_step}
\end{figure*}

\subsection{Main Results}
\label{sub:main_results}
We compare the proposed method, referred to as Time-Channel Distillation (TCD), with existing class-incremental learning baselines, which are originally designed for the class-incremental image classification task.
Especially, we choose the algorithms utilizing knowledge distillation as ours, including LwFMC~\cite{li2017learning, rebuffi2017icarl}, LwM~\cite{dhar2019learning}, iCaRL~\cite{rebuffi2017icarl}, UCIR~\cite{hou2019learning}, PODNet~\cite{douillard2020podnet}. 
We do not report the NME results from LwFMC~\cite{li2017learning, rebuffi2017icarl} and LwM~\cite{dhar2019learning} since they do not use exemplars.
To provide an upper-bound performance of our task, we introduce an oracle model, which is incrementally trained the model while preserving all the training data in the previous steps.
We reimplement the baseline algorithms and train their models using our datasets for fair comparisons.
The implementation details for the baselines are presented in the supplementary material. 

Table~\ref{tab:main_table} presents the overall results of the proposed algorithm and other baselines on UCF101 and HMDB51 datasets, where our approach outperforms all competing methods in all the experimental settings.
As mentioned earlier, NME scores for the methods without exemplars are not reported while iCaRL has the results only because the method is originally designed for the classifier.
As can be easily expected, the approaches do not exploit exemplars demonstrate poor performance. 

Table~\ref{tab:sthsth_table} presents the overall results of the proposed algorithm and recent methods~\cite{douillard2020podnet,hou2019learning} on Something-Something V2.
TCD is also more effective than other methods on the large-scale motion-sensitive dataset.
It is noticeable that, for Something-Something V2, the performance for NME falls behind CNN.
Since Something-Something V2 needs more temporal reasoning, the strategies relying on na\"ive averaging of the features from all frames may not be suitable.

Figure~\ref{fig:acc_step} presents the average accuracy over seen classes at each incremental step.
In most of the incremental steps, TCD achieves higher accuracy, which implies its great capacity to preserve the learned knowledge in the past.
Note that even though the average incremental accuracy gain on HMDB51 with 25 stages is small, the accuracy at the last incremental step is better than that of PODNet.

\subsection{Ablation Study and Analysis}
\label{sub:ablation}
We perform several ablation studies on UCF101 with 10 steps to analyze the effectiveness of our approach.

\vspace{-0.2cm}
\paragraph{Effect of each component}
\label{sub:component}
To show the effectiveness of the time-channel importance map and the frame-wise orthogonality, we conduct the experiment for variant types of our objective function, $\mathcal{L}_\text{final}^k$.
To this end, we first define the distillation loss without importance maps, which is given by
\begin{equation}
\mathcal{L}_\text{dist}'^{k}=\sum_{l=1}^{L}\sum_{t=1}^{T}\sum_{c=1}^{C_l}\|\mat{F}_{k, t,c}^l - \mat{F}_{k-1, t,c}^l \|_F^2.
\end{equation}
Table~\ref{tab:ab_loss} presents the results from several different combinations of loss terms.
The results show that all of the introduced components contribute to the performance and their combination leads to the best performance.
One noticeable thing is that applying $\mathcal{L}_\text{ortho}^k$ without $\hat{\mat{M}}_{k,t,c}^l$ also improves the performance, where the loss alleviates the correlation between the representations across frames and help the model to address feature drift issue.
\begin{table}[!t]
	\caption{Ablations study results about the objective function. 
	We demonstrate the effectiveness of the time-channel channel importance $\hat{\mat{M}}_{k,t,c}^l$ and the orthogonality among frames $\mathcal{L}_\text{ortho}^k$.
	Note that $\mathcal{L}_\text{dist}'^k$ denotes  $\mathcal{L}_\text{dist}^k$ without importance map weights, $\hat{\mat{M}}_{k,t,c}^l$.
	}
	\vspace{-0.3cm}
	\label{tab:ab_loss}
	\begin{center}
		\scalebox{0.9}{
			\renewcommand{\arraystretch}{1.25}
			\setlength\tabcolsep{7pt}
			\begin{tabular}{l|cc}
			          Objective function                                                           &  \multicolumn{1}{c}{CNN}    & \multicolumn{1}{c}{NME}  \\ \hline
				  $\mathcal{L}_\text{cls}^k$ + $\mathcal{L}_\text{dist}'^k$                                                       & 71.21 &  73.24       \\
				  $\mathcal{L}_\text{cls}^k$ + $\mathcal{L}_\text{dist}'^k$ + $\mathcal{L}_\text{ortho}^k$               &  72.31 &  74.42       \\
				  $\mathcal{L}_\text{cls}^k$ + $\mathcal{L}_\text{dist}^k$                                             &  72.61  &  74.81      \\  \hline
				  $\mathcal{L}_\text{cls}^k$ + $\mathcal{L}_\text{dist}^k$ + $\mathcal{L}_\text{ortho}^k$ (Ours)       & \textbf{73.43} &  \textbf{75.35}         \\ \hline
			\end{tabular} 
		}
	\end{center}
	\vspace{-0.5cm}
\end{table}

\begin{table*}[!ht]
	\caption{Analysis about the memory budget for each class on UCF101 with 10 steps.
	The results show the robustness of our algorithm to varying memory budgets.}
	\vspace{-0.6cm}
	\label{tab:ab_memory}
	\begin{center}
		\scalebox{0.9}{
			\renewcommand{\arraystretch}{1.0}
			\setlength\tabcolsep{7pt}
			\begin{tabular}{l|cccccccc}
				  \multicolumn{1}{r|}{Memory per class}                &  \multicolumn{2}{c}{1}  & \multicolumn{2}{c}{2} & \multicolumn{2}{c}{\textbf{5}} & \multicolumn{2}{c}{10}   \\ 
                                                                                                                     &   CNN  & NME      & CNN         & NME       & CNN        & NME       & CNN       & NME    \\ \hline
				 iCaRL~\cite{rebuffi2017icarl}                                    &   ---  &  58.05   &   ---      &  60.50     &  ---      &  64.51     &  ---    &    66.94    \\
				 UCIR~\cite{hou2019learning}                                   &   61.92  & 65.52    &  66.43       &   67.58    &  70.42      &  70.50     &  72.47    &  71.69        \\
				 PODNet~\cite{douillard2020podnet}                        &   63.18  &  70.96    &  65.93      &   72.78     &  71.58      &  73.75    &  75.44    &   76.39      \\ \hline 
				 TCD (Ours)                                                                         &    \textbf{64.52} & \textbf{71.96} &  \textbf{68.40} &  \textbf{73.30}    &  \textbf{73.43}   &  \textbf{75.35}         & \textbf{76.66} &\textbf{77.09}             \\ \hline
			\end{tabular} 
		}
	\end{center}
	\vspace{-0.2cm}
\end{table*}
\begin{table*}[!ht]
	\caption{Analysis about the sampling strategies for storing videos in exemplar set.
	``All" denotes the strategy to store the entire video in the exemplar memory and sample examples randomly for training.
	``Random" and ``Uniform" mean the strategies that sample frames randomly and with a equal time interval, respectively. 
	The results show that storing all frames in a video does not necessarily delivers performance improvement.}
	\vspace{-0.6cm}
	\label{tab:ab_sampling}
	\begin{center}
		\scalebox{0.9}{
			\renewcommand{\arraystretch}{1.25}
			\setlength\tabcolsep{7pt}
			\begin{tabular}{l|cccccccc}
				  \multicolumn{1}{r|}{Sampling strategy}                &  \multicolumn{2}{c}{All}  & \multicolumn{2}{c}{Random} & \multicolumn{2}{c}{Uniform} \\ 
                                                                                                                     &   CNN  & NME      & CNN         & NME       & CNN        & NME        \\ \hline
				 iCaRL~\cite{rebuffi2017icarl}                                    &   ---  &    64.33  &  ---       &      64.68 &  ---      &  64.51        \\
				 UCIR~\cite{hou2019learning}                                   &   70.22  & 70.41    &  70.38       &   70.12    &  70.42      &  70.50          \\
				 PODNet~\cite{douillard2020podnet}                        &   71.76  &  73.50    &  72.37      &   73.87     &  71.58      &  73.75        \\ \hline 
				 TCD (Ours)                                                                         &    \textbf{73.89} & \textbf{75.51} &  \textbf{73.17} &  \textbf{75.30}    &  \textbf{73.43}   &  \textbf{75.35}         \\ \hline
			\end{tabular} 
		}
	\end{center}
	\vspace{-0.4cm}
\end{table*}

\vspace{-0.2cm}
\paragraph{Effect of memory size}
\label{sub:memory}
To demonstrate the robustness of TCD with respect to the memory budget, we evaluate the performance of the compared methods by varying the memory budgets.
Table~\ref{tab:ab_memory} shows that TCD outperforms other baselines regardless of the memory budget.

\vspace{-0.2cm}
\paragraph{Sampling strategy}
\label{sub:sampling}
As discussed in Section~\ref{sec:method_exemplar}, the memory requirement for video exemplars is further reduced by storing a subset of frames in a video instead of the whole video.
We conduct the experiment to show the performance variation of class-incremental learning depending on the exemplar selection strategy.
We test the following three options: storing a whole video, sampling frames randomly, and selecting frames with a uniform time interval.
In the setting that the whole video is stored, TSM selects a predefined number of frames randomly from each exemplar at each iteration.
For the random and uniform sampling strategies, we store $T$ frames, where $T$ is given by the hyperparameter of TSM network~\cite{lin2019tsm}.

Table~\ref{tab:ab_sampling} demonstrates that the simple sampling strategies are as good as the methods with the whole videos in all the tested algorithms. 
This result implies that the diversity of sampled frames affects the overall performance marginally. 
In the context of class-incremental learning, a small subset of frames in exemplar videos are sufficient to maintain the knowledge about the corresponding video, which is a desirable property to continual learning. 
However, this experiment is limited in another aspect because our backbone model, TSM, relies only on a small number of frames.

\begin{figure}[t]
	\centering
	\includegraphics[width=1.0\linewidth]{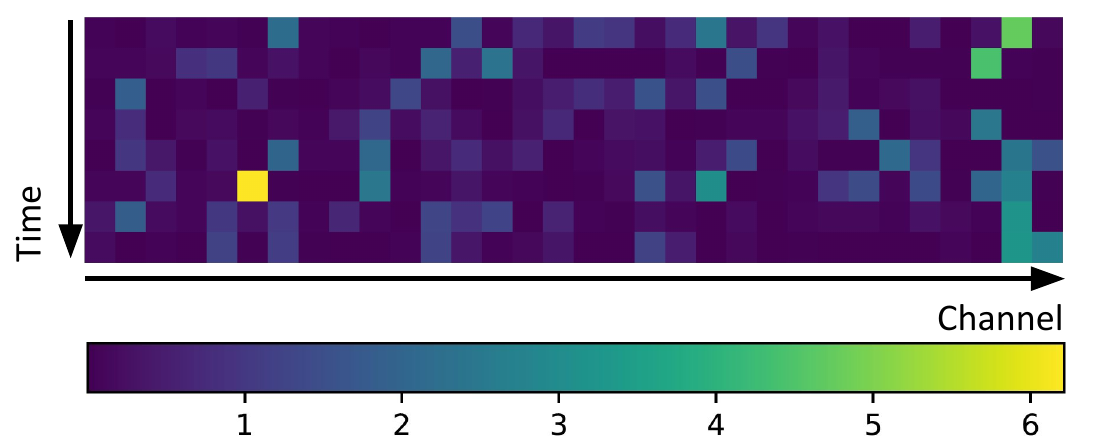}
	\caption{Visualization of the importance map obtained from the 4$^\text{th}$ ResBlock in the model trained on UCF101 with 10 stages.
	The colorbar indicates the magnitude of the estimated importance.}
	\vspace{-0.2cm}
\label{fig:importance_map}
\end{figure}

\vspace{-0.2cm}
\paragraph{Visualization of importance map}
\label{sub:vis_imp}
Figure~\ref{fig:importance_map} illustrates an example of generated importance map after the last stage training of UCF101.
The importance map for the first 32 channels of the 4$^\text{th}$ ResBlock for TSM is depicted, where the bright pixels indicate higher importance.
From the figure, one can notice that the importance of each channel varies over time.
The estimated mask makes the model leave critical features unaffected by knowledge distillation while providing the model with the flexibility to update unimportant features.

\section{Conclusion}\label{sec:conclusion}	
We presented a novel framework for class-incremental learning in the context of video action recognition, which has not been actively investigated yet.
Specifically, we introduced a new knowledge distillation loss based on time-channel importance masks, which aims to preserve crucial feature maps for preventing the catastrophic forgetting problem and make trivial ones flexible for absorbing new knowledge.
To effectively exploit the proposed distillation loss, we add a regularization term, which encourages individual feature maps along the time axis to be orthogonal to each other.
Our algorithm achieves outstanding performance compared to existing image-specific class-incremental learning approaches on multiple standard datasets, which are newly introduced to fit class-incremental learning for videos.

\section*{Acknowledgments}
This work was partly supported by Samsung Electronics Co., Ltd., the IITP grants, and the Bio \& Medical Technology Development Program of the National Research Foundation (NRF) funded by the Korea government (MSIT) [2017-0-01779, 2017-0-01780, 2021M3A9E4080782].

\clearpage

\setcounter{section}{0}
\renewcommand{\thesection}{S\arabic{section}}

\onecolumn

\section{Relation about Parameter Regularization Methods}
\label{sub:discussion}
Our approach has something common with  some parameter regularization techniques---Elastic Weight Consolidation (EWC)~\cite{kirkpatrick2017overcoming} and Synaptic Intelligence (SI)~\cite{zenke2017continual}---in the sense that they propose weighting schemes in continual learning scenarios.
Both EWC and SI attempt to regularize model parameters using the weights given by either first-order information or cumulative trajectories in the parameter space, respectively; they desire to learn proper representations by backpropagation.
On the other hand, our algorithm is based on knowledge distillation, where the learned representations are regularized using the weights given by the impact of individual activation changes with respect to the final loss, and expects the model parameters to be learned for generating the desirable features.
Contrary to EWC and SI, which attempt to preserve the representations of old tasks indirectly via parameter regularization, our method optimizes the representation directly, which would be more effective for class-incremental continual learning.

\section{Comparison with Parameter Regularization Method}
\label{sec:param_reg}
To demonstrate the performance of our approach compared to parameter regularization methods, which we discuss in Section 3.6 of the main paper, we present the results from Elastic Weight Consolidation (EWC)~\cite{kirkpatrick2017overcoming} and na\"ive fine-tuning (FT) under the same memory constraint with ours in Table~\ref{tab:ewc_table}.
Although both EWC and our approach attempt to maintain important information in the previous tasks, the proposed method optimizes the objective function directly via knowledge distillation and achieves superior performance.
Note that a similar discussion has been made in~\cite{hsu2018re, van2019three} as well.

\begin{table*}[ht!]
	\caption{Class-incremental action recognition performance evaluation on UCF101 and HMDB51 between fine-tuning (FT), EWC and the proposed method.
	Note that ``E" indicates the existence of exemplars.
	The bold-faced number means the best performance.
	EWC slightly outperforms the fine-tuning, while our approach surpasses both methods by large margins.}
	\label{tab:ewc_table}
	\begin{center}
		\scalebox{1.0}{
			\renewcommand{\arraystretch}{1.2}
			\setlength\tabcolsep{7pt}
			\begin{tabular}{l|ccc|cc}
& \multicolumn{3}{c|}{UCF101} & \multicolumn{2}{c}{HMDB51} \\  \hline			 
\multicolumn{1}{l|}{Num. of classes}              &  \multicolumn{1}{c}{10 $\times$ 5 stages}   &  \multicolumn{1}{c}{5 $\times$ 10 stages} &  \multicolumn{1}{c|}{2 $\times$ 25 stages}   & \multicolumn{1}{c}{5 $\times$ 5 stages} &  \multicolumn{1}{c}{1 $\times$ 25 stages}   \\ \hline
				 FT + E                                                               &      67.65        &    66.67         &     65.36           &    38.58      &   34.83     \\
				 EWC~\cite{kirkpatrick2017overcoming} + E               &      69.70       &     68.12      &    67.00       & 39.98     &   35.94    \\ \hline  
				 TCD (Ours) w/o $\mathcal{L}_\text{ortho}$   &    \textbf{73.09}      &\textbf{72.61} & \textbf{71.33} &  \textbf{45.14} &  \textbf{46.11}    \\ 
				 TCD (Ours)                       &    \textbf{74.89}      &\textbf{73.43} & \textbf{72.19} &  \textbf{45.34} &  \textbf{46.66}          \\ \hline			\end{tabular} 
		}
	\end{center}
\end{table*}
\vspace{-0.7cm}

\section{Compatibility with Bias Correction Method}
\label{sec:bias_cor}
In order to show the compatibility of our distillation objective to other kinds of algorithms, we combine the proposed method with an existing bias correction method, Bias Correction (BiC)~\cite{wu2019large}.
For a fair comparison, we replace the classifier of our model with a linear classifier.
We set the ratio between training/validation split on the exemplars to $4:1$ to perform BiC method, as we use 5 exemplars per class.
Table~\ref{tab:bic_table} illustrates the results on UCF101.
The performance gap between BiC combined with ours and BiC become larger when the number of incremental steps increase, which implies the robustness of the proposed approach.

\begin{table*}[ht!]
	\caption{Compatibility of our distillation loss with the bias correction (BiC) method on UCF101.
	The bold-faced number indicates the best performance.}
	\label{tab:bic_table}
	\begin{center}
		\scalebox{1.0}{
			\renewcommand{\arraystretch}{1.2}
			\setlength\tabcolsep{7pt}
			\begin{tabular}{l|ccc}		 
\multicolumn{1}{l|}{Num. of classes}              &  \multicolumn{1}{c}{10 $\times$ 5 stages}   &  \multicolumn{1}{c}{5 $\times$ 10 stages} &  \multicolumn{1}{c}{2 $\times$ 25 stages}    \\ \hline
				 BiC                                                               &     77.00        &    74.94         &     68.85      \\
				 BiC + TCD (Ours)                      &    \textbf{77.22}      &\textbf{75.63} & \textbf{72.00}        \\ \hline			\end{tabular} 
		}
	\end{center}
\end{table*}

\section{Effect of Number of Input Frames}
We set 8 frames as the input size following the convention of action recognition models~\cite{lin2019tsm,tran2019video}.
However, our algorithm also works well with different input sizes, which incur a trade-off between accuracy and  cost.
Table~\ref{tab:frame_table} demonstrates the results by varying the number of frames on UCF101 with the same backbone model, TSM.
It shows that our algorithm consistently outperforms PODNet regardless of the number of input frames.


\begin{table*}[h!]
	\caption{Effect of input size on UCF101 with 10 stages.} 
	\label{tab:frame_table}
	\begin{center}
	\vspace{0.3cm}
	\scalebox{1.0}{
			\setlength\tabcolsep{4pt}
			\begin{tabular}{l|cc|cc|cc|cc}		
\multicolumn{1}{l|}{\# of frames}              &  \multicolumn{2}{c|}{16}   &  \multicolumn{2}{c|}{8}  &  \multicolumn{2}{c|}{4} &  \multicolumn{2}{c}{1}    \\ 
Classifier &   CNN  & NME      & CNN         & NME  &   CNN  & NME      & CNN         & NME      \\ \hline \hline
				 PODNet                     & 73.36  & 74.80  &   71.58  &  73.75   &  70.93     &   73.47    &  68.89    &  71.98    \\ 
				 TCD (Ours)                        &   \textbf{75.17}  & \textbf{76.13}       & \textbf{73.43}             & \textbf{75.35}              & \textbf{71.17}  & \textbf{74.18}   &  \textbf{69.14}    &  \textbf{72.93} \\ \hline
			\end{tabular} 
		}
		\end{center}
\end{table*}


\section{Implementation Details}
\label{sec:impl_detail}
For all experiment, we set the initial learning rate as 0.001 and adopt the SGD optimizer with weight decay of 0.0005.
The learning rate is divided by 10 after 20 and 30 epochs.
We construct our Local Similarity Classifier (LSC) by using 3 proxies and allow $\eta$ to be trained throughout the training procedure. 
For UCF101, we set $\alpha=1.0$ for the intermediate features and $0.01$ for the logit, and set $\beta=0.1$ for $\mathcal{L}_\text{ortho}$.
For HMDB51, we set $\alpha=3.0$ for the intermediate features and $0.1$ for the logit, and set $\beta=0.3$.
For Something-Something V2, we set $\alpha=0.5$ for the intermediate features and $10.0$ for the logit, and set $\beta=10^{-3}$.
Following PODNet~\cite{douillard2020podnet} and UCIR~\cite{hou2019learning}, we further multiply an adaptive scaling factor $\lambda=\sqrt{\frac{| \mathcal{C}_{1:k} |}{|\mathcal{C}_k |}}$ to $\alpha$ at each incremental step $k$, where $ \mathcal{C}_{1:k}=\mathcal{C}_1 \cup \cdots \cup \mathcal{C}_{k}$ denotes the number of class observed until the incremental step $k$.

To compare our method to the existing continual learning methods, we reimplement each algorithm and search the hyperparameters using a grid search.
For UCIR~\cite{hou2019learning} and PODNet~\cite{douillard2020podnet},  we explore the hyperparameters for the distillation losses, $a\cdot 10^{b}$, with $a \in \{1,3,5\}$ and $b\in \{-2,\cdots,2\}$.
As a result, we set the weight for UCIR as $5$ and set the rest of the hyperparameters as same as the original paper.
PODNet has two distillation terms, the distillation term for intermediate features and the logit.
We set the weight of each loss term to $(0.5, 3.0)$, $(0.1, 1.0)$, and $(1.0, 5.0)$ for UCF101, HMDB51, and Something-Something V2 respectively.
The margin for the LSC is set to 0.6 for both PODNet and ours.

\twocolumn

{\small
\bibliographystyle{ieee_fullname}
\bibliography{egbib}
}

\end{document}